\documentclass[conference, letterpaper]{IEEEtran}
\IEEEoverridecommandlockouts

\usepackage{cite}

\ifCLASSINFOpdf
  \usepackage[pdftex]{graphicx}
  \DeclareGraphicsExtensions{.pdf,.jpeg,.png,.eps}
\else
\fi
\usepackage[cmex10]{amsmath}
\usepackage{algpseudocode}
\usepackage[tight,footnotesize]{subfigure}
\usepackage{bbm}
\usepackage{amssymb}

\newtheorem{prop}{Proposition}

\newtheorem{remark}{Remark}

\DeclareMathOperator*{\argmin}{\arg\!\min}

\usepackage{color}
\usepackage[textwidth=1.850cm, textsize=scriptsize]{todonotes}

\begin{document}
%
\title{Calibration of One-Class SVM for MV set estimation}




\author{
\IEEEauthorblockN{Albert Thomas \IEEEauthorrefmark{1}\IEEEauthorrefmark{2},
Vincent Feuillard \IEEEauthorrefmark{1} and
Alexandre Gramfort \IEEEauthorrefmark{2}}
\IEEEauthorblockA{\IEEEauthorrefmark{1}Airbus Group Innovations, first.last@airbus.com, 12 rue Pasteur, 92150, Suresnes, France}
\IEEEauthorblockA{\IEEEauthorrefmark{2}LTCI, CNRS, T\'el\'ecom Paris-Tech, Universit\'e Paris-Saclay, first.last@telecom-paristech.fr, 75013, Paris, France}
\thanks{Copyright notice: 978-1-4673-8273-1/15/\$31.00 \copyright 2015 European Union}
}


\maketitle

\begin{abstract}
A general approach for anomaly detection or novelty detection consists in estimating high density regions or Minimum Volume (MV) sets. The One-Class Support Vector Machine (OCSVM) is a state-of-the-art algorithm for estimating such regions from high dimensional data. Yet it suffers from practical limitations. When applied to a limited number of samples it can lead to poor performance even when picking the best hyperparameters. Moreover the solution of OCSVM is very sensitive to the selection of hyperparameters which makes it hard to optimize in an unsupervised setting. We present a new approach to estimate MV sets using the OCSVM with a different choice of the parameter controlling the proportion of outliers. The solution function of the OCSVM is learnt on a training set and the desired probability mass is obtained by adjusting the offset on a test set to prevent overfitting. Models learnt on different train/test splits are then aggregated to reduce the variance induced by such random splits. Our approach makes it possible to tune the hyperparameters automatically and obtain nested set estimates. Experimental results show that our approach outperforms the standard OCSVM formulation while suffering less from the curse of dimensionality than kernel density estimates.
Results on actual data sets are also presented.
\end{abstract}


%
\IEEEpeerreviewmaketitle

\section{Introduction}

An anomaly is defined as any observation that does not conform to the expected normal behavior \cite{Chandola2009}. The goal of anomaly detection also referred as novelty detection is to identify abnormal observations without previously knowing them. Applications include machine fault detection, network intrusion detection in cybersecurity or fraud detection in finance.
Given observations $X_1, \dots, X_n \in \mathbb{R}^d$, $d \geq 1$, independent and identically distributed realizations of an unknown probability distribution $P$, we would like to learn a subset of $\mathbb{R}^d$ such that points lying inside this set will be considered as normal and points lying outside will be considered as anomalies. The implicit hypothesis made in this context is that anomalies correspond to rare events and are located in the tail of the distribution. A possible approach is to estimate the subset corresponding to the region where the data are most concentrated. Such a region is called a Minimum Volume (MV) set, i.e., the set of minimum volume with probability mass at least $\alpha$, with $\alpha$ close to 1.


The notion of MV sets has been introduced by Polonik \cite{Polonik1997}. Let $\mu$ be the Lebesgue measure and $\alpha \in (0,1)$. A MV set with mass at least $\alpha$ is a solution of the following optimization problem
\begin{equation}
\label{MVsetpb}
\min_{G \in \mathcal{B}(\mathbb{R}^d)} \mu(G) \quad \text{such that } P(G) \geq \alpha \enspace ,
\end{equation}
where $\mathcal{B}(\mathbb{R}^d)$ is the set of all measurable subsets of $\mathbb{R}^d$.

We assume that the probability measure $P$ has a density $h$ with respect to the Lebesgue measure $\mu$ and that $h$ has no flat parts, i.e., $\mu(\{x, h(x)=\tau\}) = 0$ for all $\tau > 0$. One can show that under regularity assumptions on $h$, the optimization problem \eqref{MVsetpb} has an unique solution $G^*_{\alpha}$ (up to subsets of null $\mu$-measure). This solution satisfies $P(G^*_{\alpha}) = \alpha$ and is a density level set, i.e., a set of the form $\{h \geqslant \tau \}, \tau > 0$ \cite{Einmahl1992}. A MV set is thus a density level set. The converse holds with no assumption on the density: density level sets are MV sets.

There are essentially two different approaches to estimate a MV set. The first one is to resort to a plug-in approach where one first estimates the underlying density and then thresholds it at the level $\hat \tau_{\alpha}$ such that $P(\{ \hat h_n \geq \hat \tau_{\alpha} \}) = \alpha$ where $\hat h_n$ is a density estimator. The main drawback of this approach is that plug-in estimators do not scale well with the dimension (for e.g. see \cite{Baillo2003, Cadre2006, Cadre2013}). Moreover the entire density is estimated while just a density level set is needed.

The second one is to resort to a direct approach by choosing the set of minimum volume containing a proportion $\alpha$ of the sample points among a class of sets such as Glivenko-Cantelli or Vapnik-Cervonenkis classes. Direct approach algorithms include algorithms from \cite{Scott2006, Davenport2006a} and the OCSVM \cite{Scholkopf2001, Tax2004}. Scott and Nowak \cite{Scott2006} introduce a framework analogous to the empirical risk minimization in binary classification to estimate a MV set. Davenport et al. \cite {Davenport2006a} use a Neyman-Pearson classification approach to estimate MV sets with SVMs or any other classification algorithms. Tax and Duin \cite{Tax2004} introduce the Support Vector Data Description (SVDD) algorithm to search for the hypersphere with the minimum volume containing at least a proportion $\alpha$ of data sample in a Reproducing Kernel Hilbert Space (RKHS). If the kernel used is the Gaussian kernel then the OCSVM and SVDD are equivalent \cite{Tax2001a}.

While the problem of anomaly detection is unsupervised, it is known that an unsupervised problem can be transformed into a supervised one \cite{Hastie2009}. Steinwart et al. \cite{Steinwart2005} introduce a classification framework for density level set estimation. The classification is performed between the original data and an artificial second class. The density level set $\{h \geqslant \tau_{\alpha} \}$ can then be learnt with any classification algorithm without estimating the entire density $h$. However one still needs to choose the threshold corresponding to a mass $\alpha$ which can be computationally expensive.


The OCSVM algorithm introduced by Sch\"{o}lkopf et al. \cite{Scholkopf2001} is one of the most popular algorithm for anomaly and novelty detection. In \cite{Vert2006}, Vert and Vert show that the OCSVM is a consistent estimator of density level sets. In fact they give a more powerful result: the solution function returned by the OCSVM gives an estimate of the tail of the underlying density $h$. The OCSVM is mainly applied with the Gaussian kernel and the performance highly depends on the kernel bandwidth selection.

\begin{figure}[!t]
\centering
\includegraphics[width=3.6in]{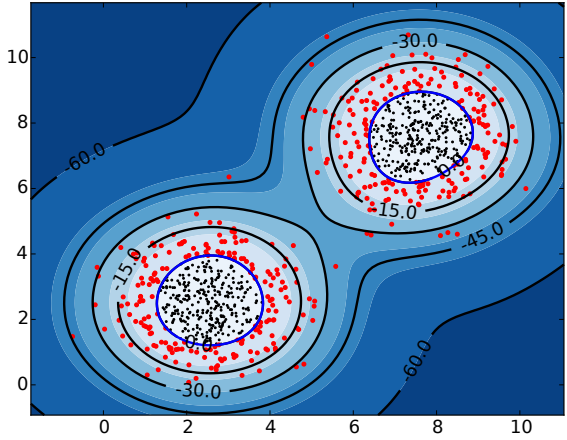}
\caption{Application of the OCSVM with $\nu = 0.4$ on a Gaussian mixture sample of size $n=1000$. In blue the estimated set, in black the level sets of the solution function of the OCSVM, in red the support vectors. The solution function captures the structure of the tail of Gaussian mixture distribution.}
\label{gm_1000_95_nu4}
\end{figure}

With the formulation introduced by Sch\"{o}lkopf et al. \cite{Scholkopf2001}, the mass of the estimated set is controlled by a parameter $\nu$ specified by the user. The estimated set is guaranteed to contain at least a fraction $1 - \nu$ of the data. However simple simulations show that the OCSVM can perform very poorly to estimate a MV set for a finite data sample. For instance for a Gaussian mixture such as the one in Figure \ref{gm_1000_95_nu4}, no value of the kernel bandwidth gives a good approximation of the true MV set with mass at least 0.95 when the parameter $\nu$ is chosen such that the empirical probability of the estimated set is larger than $\alpha$ (see section \ref{experiments_gm}). However using a different value of $\nu$, the set estimated by the OCSVM clearly differs from the true MV set but the solution function captures the structure of the tail of the underlying distribution as shown in Figure \ref{gm_1000_95_nu4}.

The approach we propose and describe in the second part of this paper consists in fixing $\nu$ at a value such that the proportion of points outside the estimated set will be strictly greater than $1-\alpha$. The solution function is learnt on a training set and then thresholded to obtain the desired probability mass on a test set to prevent overfitting. To reduce the variance induced by the random split of the data set into a training set and a test set we aggregate several models.
Thresholding the solution function of the OCSVM to obtain the desired probability mass is an approach that has already been very briefly mentionned in \cite{Scholkopf2001} and in \cite{Martinez2008}. However, to the best of our knowledge, such an approach has never been considered thoroughly. 
In the second part of this paper we present the OCSVM and its properties before presenting our approach. In the last part we compare the performance of our approach with the OCSVM on simulated data sets and apply our approach to real data sets. Connections can be made between this paper and \cite{Filippone2010} in which Filipone et al. apply the possibilistic $c$-means algorithm in kernel-induced spaces.


%

\section{Method}
\label{OCSVMsection}

\subsection{Background on One-Class SVM} 
The OCSVM was introduced by Sch\"{o}lkopf et al. \cite{Scholkopf2001} to estimate high density regions from a data sample. After mapping the data in a feature space through a function $\Phi$ determined by a specific kernel $k$ the OCSVM finds a separating hyperplane between the origin and the mapped data. The separating hyperplane defined by a vector $w$ and an offset $\rho$ is given by the solution of the following optimization problem
\begin{equation}
\label{nuPrimal}
\begin{aligned}
\min_{w,\xi,\rho} \quad &\frac{1}{2}\Vert w \Vert^2 - \rho + \frac{1}{\nu n}\sum_{i=1}^n\xi_i\\
\text{s.t.} \quad & \langle w, \Phi(x_i) \rangle \geq \rho - \xi_i \quad, \quad 1 \leq i \leq n \\
& \xi_i \geq 0 \quad, \quad 1 \leq i \leq n 
\end{aligned}
\end{equation}
where $\nu \in (0,1)$ is a parameter specified by the user. This problem is convex and as strong duality holds it is solved through its dual
\begin{equation}
\label{nuDual}
	\begin{aligned}
	\min_{\gamma} \quad &\frac{1}{2}\sum_{1 \leq i,j \leq n}{\gamma_i \gamma_j k(x_i,x_j)}\\
	\text{s.t.} \quad & 0 \leq \gamma_i \leq \frac{1}{\nu n} \quad, \quad 1 \leq i \leq n \\
	& \sum_{i=1}^n{\gamma_i} = 1 
	\end{aligned}
\end{equation}

The resulting solution function is given by
\begin{equation*}
x \mapsto \sum_{i=1}^n \gamma_i k(x, x_i)
\end{equation*}
and the resulting estimated MV set by
\begin{equation}
\label{OCSVMset}
\hat G = \{ x, \sum_{i=1}^n \gamma_i k(x, x_i) - \rho_{\nu} \geq 0\}
\end{equation}
where $\rho_{\nu}$ denotes the $\rho$ solution of \eqref{nuPrimal}.

As with SVM in supervised settings, not all the $\gamma_i$ are non-zero. The points $x_i$ such that $\gamma_i > 0$ are called support vectors (SVs). Support vectors are exactly the samples located outside or on the border of the set $\hat G$:
\begin{equation*}
\{x_j, 1 \leq j \leq n, \sum_{i=1}^n \gamma_i k(x_j, x_i) - \rho_{\nu} \leq 0 \} \enspace .
\end{equation*}

Outliers are exactly the samples that are located strictly outside the set $\hat G$:
\begin{equation*}
\{ x_j, 1 \leq j \leq n, \sum_{i=1}^n \gamma_i k(x_j, x_i) - \rho_{\nu} < 0 \} \enspace .
\end{equation*}

The parameter $\nu$ needs to be chosen by the user. We have the following property \cite{Scholkopf2001}: \newline

\begin{prop}
\label{nuproposition}
Assuming the solution of \eqref{nuPrimal} satisfies $\rho_{\nu} > 0$ the following statement holds
\begin{itemize}
\item[i)] $\nu$ is an upper bound on the fraction of outliers and a lower bound on the fraction of SVs
\begin{equation*}
\frac{\text{Outliers}}{n} \leq \nu \leq \frac{\text{SV}}{n} \enspace .
\end{equation*}
\item[ii)] If the data were generated independently from a distribution $P$ absolutely continuous with respect to the Lebesgue measure and if the kernel $k$ is analytic and non constant then
\begin{equation*}
\begin{aligned}
\frac{\text{SV}}{n} \longrightarrow \nu \quad \text{almost surely} \\
\frac{\text{Outliers}}{n} \longrightarrow \nu \quad \text{almost surely}
\end{aligned}
\end{equation*}
\end{itemize}
\end{prop}

This property is of great interest in practice. It gives the user some insights on how to choose the parameter $\nu$. Indeed the empirical probability of the estimated set is greater than $1 - \nu$ and the probability of the estimated set converges almost surely to $1-\nu$ as $n$ tends to infinity. Hence one possible approach is to choose $\nu = 1 - \alpha$ to estimate a MV set with mass at least $\alpha$ \cite{Glazer2012,Glazer2013}.

In the following the kernel $k$ is the Gaussian kernel $k_{\sigma}, \sigma >0$, and is defined as
\begin{equation*}
k_{\sigma}(x, x') = \exp\left(-\frac{1}{2 \sigma^2}\Vert x - x' \Vert^2\right) \enspace .
\end{equation*}
We denote by $f_{\sigma}$ the solution function
\begin{equation*}
f_{\sigma}(x) = \sum_{i=1}^n \gamma_i k_{\sigma}(x, x_i) \enspace .
\end{equation*}

The paper of Vert and Vert \cite{Vert2006} proves the consistency of the OCSVM for density level sets estimation and hence for MV sets estimation. The optimization problem associated with the OCSVM studied in their paper is the following

\begin{equation}
\label{VertFormulation}
	\min_{f \in H_\sigma} \quad  \frac{1}{n}\sum_{i=1}^n \max(0,1-f(x_i)) + \lambda \Vert f \Vert^2_{H_\sigma}
\end{equation}
where $H_\sigma$ is the RKHS associated to the normalized Gaussian kernel and $\lambda > 0$ a regularization parameter.


Vert and Vert \cite{Vert2006} prove that for a well calibrated kernel bandwidth $\sigma$, the OCSVM is a consistent estimator of every density level sets of level $\tau \in (0, 2\lambda)$. To show such a result they prove that the solution of the OCSVM when a normalized Gaussian kernel is used converges in norm $L^2$ and in probability to the underlying density truncated at $2 \lambda$:
\begin{equation*}
\lim_{n \rightarrow + \infty} \Vert f_{\sigma} - h_{\lambda} \Vert_{L_2} = 0 \quad \text{in probability}
\end{equation*}
where 
\begin{equation*}
h_{\lambda} =
\begin{cases}
\dfrac{h(x)}{2 \lambda} \quad \text{if } h(x) \leq 2 \lambda \\
1 \quad \text{otherwise}.
\end{cases}
\end{equation*}

\begin{remark}[Connection with kernel smoothing]
If $\nu = 1$ the constraints of the dual problem $\eqref{nuDual}$ give $\gamma_i = \frac{1}{n}$ for all $i \in \{1, \dots, n\}$. This means that all the samples are taken into account in the solution and the solution function is
\begin{equation*}
f_{\sigma}(x) = \frac{1}{n}\sum_{i=1}^n k_{\sigma}(x, x_i) \enspace .
\end{equation*}
This function is the one we recover when performing a kernel smoothing with the same kernel bandwidth $\sigma$ in all the directions.

The advantage of the OCSVM over a kernel smoothing is that the estimated set is only characterized by the support vectors which, for small values of $\nu$, represent a small fraction of the sample size: the solution is sparse. This property is useful when performing the prediction task which is therefore less expensive than when using a kernel smoothing approach. Besides the solution function gives an approximation of the tail of the underlying density and, unlike a kernel smoothing, the approximation given by the solution function can be very bad elsewhere. This is why classification is sometimes said to be easier than regression \cite{Devroye1996}: we only want to be good in a neighborhood of the border of the set of interest and not elsewhere. \newline
\end{remark}


Eventually, parametrization of the mass of the MV set estimated by the OCSVM via the parameter $\nu$ does not allow to obtain nested set estimates as the mass $\alpha$ increases. For each $\nu$ a new optimization problem is solved and nothing ensures that the different set estimates are nested. Variants of the OCSVM that ensure this property have been introduced \cite{Lee2007, Lee2010}. With our approach, the mass of the MV set is parametrized through the offset and this allows us to produce nested sets in a neighborhood of the estimated MV set with mass at least $\alpha$. For the same solution function, we select different offsets $\rho$, one for each mass.

\subsection{Automatic Calibration of OCSVM}
\label{approach}

We want to estimate a MV set with mass at least $\alpha$ with $\alpha$ close to $1$ from the sample $X_1, \dots, X_n$. Thanks to the result of Vert and Vert \cite{Vert2006}, we know that the solution function of the OCSVM gives an approximation of the tail of the underlying distribution. More precisely in our approach we use the fact that $f_{\sigma}$ is an approximation of the underlying density in a neighborhood of the border of the MV set. The algorithm we propose is described in Figure \ref{OCSVMoffset} and detailed hereafter.

\begin{figure}[!t]
\centering
\begin{algorithmic}
  \State \textbf{Input}: parameter $\nu$, mass $\alpha$, data set $X$, kernel bandwidths set $\Sigma$, $c > 0$
	\State{Randomly split $X$ in a training set $X_{train}$ and a test set $X_{test}$}
	\For{kernel bandwidth $\sigma$ in $\Sigma$}
		\State{$f_{\sigma} = \text{OCSVM}(\nu, \sigma, X_{train})$}
		\For{$\beta$ in $[\alpha - c, \alpha + c]$}
			\State Bisection search to find $\hat \rho_{\beta}$ such that
			\begin{equation*}
			P_{n_{test}}(\hat G^\sigma_{\hat \rho_{\beta}}) = \beta
			\end{equation*}
			where $\hat G^\sigma_{\hat \rho_{\beta}} = \{x, f_{\sigma}(x) - \hat \rho_{\beta} \geq 0 \}$
			\State Computation of $\mu^{\sigma}_{\hat \rho_{\beta}} = \mu(\hat G^\sigma_{\hat \rho_{\beta}})$ by Monte Carlo integration
		\EndFor
	\EndFor
	\State Compute Area under the Mass Volume curve $(\beta,\mu^{\sigma}_{\hat \rho_{\beta}})$ for each $\sigma$: AMV($\sigma$)
	\State $\sigma_{opt} = \argmin_{\sigma \in \Sigma} \text{AMV}(\sigma)$
	\State \Return $\hat G^{\sigma_{opt}}_{\hat \rho_{\alpha}} = \{x, f_{\sigma_{opt}}(x) - \hat \rho_{\alpha} \geq 0 \}$
\end{algorithmic}
\caption{Algorithm of the OCSVM with a calibrated offset and the selection of the optimal kernel bandwidth}
\label{OCSVMoffset}
\end{figure}

First the data set $X = (X_1,\dots,X_n)$ is randomly split in a training set $X_{train}$ and a test set $X_{test}$ respectively of size $n_{train}$ and $n_{test}$. Let $\hat G$ be the set estimated by the OCSVM on the training set. The parameter $\nu$ is chosen such that we are able to estimate the underlying distribution for the interval of masses $[\alpha - c, \alpha + c]$ where $c > 0$. Therefore $\nu$ must be chosen such that $P_{n_{train}}(\hat G) \leqslant \alpha -c$, where $P_{n_{train}}$ denotes the empirical probability measure based on the training set. $P_{n_{train}}(\hat G) \leqslant \alpha -c$ is equivalent to a fraction of outliers, points lying outside $\hat G$, greater than $1 - (\alpha - c)$. What we have from proposition \ref{nuproposition} is that the fraction of outliers is less than $\nu$ for all $n$ and converges almost surely to $\nu$ as $n$ tends to infinity. The closer $\nu$ is to 1, the more outliers we allow the OCSVM to find. If $\nu$ has been set such that the fraction of outliers is less than $1 - (\alpha - c)$, then a higher value should be chosen. As we only consider values of $\alpha$ close to 1, we do not need $\nu$ to be too close to $1$ and can therefore preserve the sparsity of the OCSVM. In our algorithm we assume that a good value for $\nu$ is known and is set independently of the data set.

The function $f_{\sigma}$ gives an approximation of the tail of the distribution. Consequently thresholding it at $\hat \rho_{\alpha}$ such that $P_{n_{test}}(f_{\sigma} \geq \hat \rho_{\alpha}) = \alpha$ should offer an approximation of the MV set with mass at least $\alpha$, where $P_{n_{test}}$ denotes the empirical probability measured based on the test set. \newline


\begin{remark}
Let $\alpha_1 < \dots < \alpha_N$ be $N$ values in $[\alpha - c, \alpha + c]$ and let $\hat \rho_1 \geq \dots \geq \hat \rho_N$ be such that for all $i \in \{1, \dots, N\}$ we have $P_{n_{test}}(f_{\sigma} \geq \hat \rho_i) = \alpha_i$. Let $\hat G_i$ be the set $\hat G_i = \{x, f_{\sigma}(x) \geq \hat \rho_i \}$, then by construction the following holds
\begin{equation*}
\hat G_1 \subset \dots \subset \hat G_N \enspace .
\end{equation*}
\end{remark}

%


\subsection{Performance metric and kernel bandwidth selection}

To assess the performance of our approach and select the kernel bandwidth we need a performance metric. The kernel bandwidth parameter selection is an important task in practice as the solution of OCSVM highly depends on its choice. Low values of $\sigma$ lead to overfitting. On the contrary, high values of $\sigma$ lead to underfitting.

A performance metric used for the theoretical study of MV sets or density level set estimators is the Lebesgue measure of the symmetric difference between the true MV set $G^*_{\alpha}$ and the estimate $\hat G$, $\mu(G^*_{\alpha} \Delta \hat G)$ where $A \Delta B = (A \backslash B) \cup (B \backslash A)$ \cite{Scott2006, Steinwart2005, Rigollet2009}. 

This performance metric depends on the true MV set $G^*_{\alpha}$. We use it to assess the performance of our approach and select the optimal kernel bandwidth when we have access to the true MV set.

Several performance metrics have been used to assess the quality of one-class classification algorithms and select the optimal hyperparameters (see among others \cite{Tax2001b, Lee2007, Lee2010, Davenport2006a}). It is noteworthy to say that all these metrics require to sample points uniformly, either to compute the volume of the estimated set or to generate an artificial second class. Therefore both method suffer from the curse of dimensionality. First, the proportion of points uniformly sampled in the hypercube enclosing the data lying in the estimated set can decrease exponentially to $0$ with the dimension. Second, for high dimensions, data are expected to be very sparse and to be very easily separated, leading classification solutions to overfit. We must therefore limit the use of these metrics to data sets of low dimension, for e.g. $d \leq 10$. This has been mentionned by Tax in \cite{Tax2001b, Tax2001a}.

The performance metric we decide to use in our algorithm to select the kernel bandwidth is the Mass Volume curve introduced by Cl\'emen\c{c}on and Jakubowicz \cite{Clemencon2013} and defined as $\{(\alpha,\mu(G^*_{\alpha})), \alpha \in (0,1)\}$. To use this performance metric, we still need to sample points uniformly to compute the volume. The Mass Volume curve is a functional criterion that can be used to assess the quality of a scoring rule in the unsupervised setting. The Mass Volume curve of the true underlying distribution is the lowest Mass Volume curve that can be obtained. Cl\'{e}men\c{c}on and Robbiano \cite{Clemencon2014} give the explicit relation between the well known area under the ROC curve (AUC) and the area under the Mass Volume curve. Minimizing the area under the Mass Volume curve is equivalent to maximizing the AUC when the second class has been generated from a uniform distribution.

The Mass Volume curve is suited to assess the quality of scoring rules whereas the first purpose of the OCSVM is not to estimate a scoring rule. Indeed, the OCSVM with $\nu = 1- \alpha$ gives an estimated set of the form $\{x, f_{\sigma}(x) \geq \rho_{\nu}\}$. However there is no guarantee that for all $\rho \ne \rho_{\nu}$, sets of the form $\{x, f_{\sigma}(x) \geq \rho\}$ are good approximations of MV sets. Our approach estimates a scoring rule for the points located in the tail of the distribution and we use the area under the Mass Volume curve for masses in a neighborhood of $\alpha$ as a performance metric to select the best kernel bandwidth.

To compute the Mass Volume curve, $\{(P(\hat G_{\beta}), \mu(\hat G_{\beta})), \beta \in [\alpha - c, \alpha +c]\}$, we need to compute the probability and the volume of the estimated set. The probability is estimated on the test set and is thus equal to $\beta$ as we choose the offset such that the empirical probability of the estimated set on the test set equals $\beta$. We estimate the volume by Monte Carlo estimation.\newline

\noindent
\emph{Volume computation:}
The volume of a set $G = \{x, f_{\sigma}(x) \geq \rho\}$ is defined as
\begin{equation}
\label{volume}
\mu(G) = \int \mathbbm{1}_G(x) \mu(dx) \enspace .
\end{equation}

This integral cannot be computed exactly so we resort to Monte Carlo estimation. As we do not know how to sample uniformly in the set $G$ either we resort to importance sampling rewriting \eqref{volume} as

\begin{equation}
\mu(G) = \int \frac{\mathbbm{1}_G(x)}{q(x)} q(x)\mu(dx)
\end{equation}
where $q$ must be a well chosen distribution.

The most popular distribution used in the literature is the uniform distribution over the hypercube $G_c$ enclosing the data. Let $V_c$ be the volume of $G_c$ then the density of such a distribution is $q_c(x) = \frac{1}{V_c}\mathbbm{1}_{G_c}(x)$ and

\begin{equation*}
\begin{aligned}
\mu(G) &= V_c\int \frac{\mathbbm{1}_G(x)}{\mathbbm{1}_{G_c}(x)} q_c(x)\mu(dx) = V_c \int \mathbbm{1}_G(x) q_c(x)\mu(dx) \\
&= V_c \mathbb{E}_{q_c}[\mathbbm{1}_G(Z)] \enspace .
\end{aligned}
\end{equation*}
Thanks to the Law of Large Numbers the volume $\mu(G)$ is estimated by

\begin{equation*}
\label{volumecubeest}
\hat \mu_c(G) = \frac{V_c}{m}\sum_{i=1}^m \mathbbm{1}_G(Z_i) \quad \quad Z_i \sim q_c .
\end{equation*}


%
%

Sampling uniform data is an issue worth mentioning as it is the factor limiting the estimation of Minimum Volume sets in a high dimension setting. \newline

\subsection{Aggregation}

In section \ref{approach} we presented our approach consisting in the following:
\begin{enumerate}
\item Randomly split the data set in a training set and a test set
\item Train the OCSVM on the training set to obtain $f_{\sigma}$
\item Find the offset $\hat \rho_{\alpha}$ such that $P_{n_{test}}(\{f_{\sigma} \geq \hat \rho_{\alpha}\}) = \alpha$ on the test set
\end{enumerate}

Randomly splitting the data set in training and test sets introduces variance in the result. To reduce the variance we aggregate several models based on $B$ train/test splits. Let $(f^b_{\sigma}, \hat \rho^b_{\alpha})$, $1 \leq b \leq B$ be the models obtained, where $\hat \rho^b_{\alpha}$ is such that
\begin{equation*}
P^b_{n_{test}}(\{x, f_{\sigma}(x)- \hat \rho^b_{\alpha} \geq 0 \}) = \alpha \enspace .
\end{equation*}

Averaging all the models we obtain
\begin{equation*}
F^B_{\sigma}(x) = \frac{1}{B}\sum_{b=1}^B (f^b_{\sigma}(x) - \hat \rho^b_{\alpha}) \enspace .
\end{equation*}

The final estimated set is given by

\begin{equation*}
\hat G^B_{\alpha} = \{x, F^B_{\sigma}(x) \geq 0\} \enspace .
\end{equation*}

The algorithm is described in Figure \ref{OCSVMoffsetVariant}. \newline


\begin{figure}[t!]
\centering
\begin{algorithmic}
  \State{\textbf{Input}: parameter $\nu$, mass $\alpha$, data set $X$, kernel bandwidths set $\Sigma$, $c>0$, number of models $B$}
	\For{$b$ in $\{1, \dots, B \}$}
    \State Randomly split $X$ in a training set $X_{train}$ and a test set $X_{test}$
    \For{kernel bandwidth $\sigma$ in $\Sigma$}
		  \State $f^b_{\sigma} = \text{OCSVM}(\nu, \sigma, X_{train})$
      \For{$\beta$ in $[\alpha -c, \alpha +c]$}
		    \State Bisection search to find $\hat \rho^b_{\beta}$ such that 
		    \begin{equation*}
		    P_{n_{test}}(\hat G^{\sigma}_{\hat \rho^b_{\beta}}) = \beta
		    \end{equation*}
        where $\hat G^{\sigma}_{\hat \rho^b_{\beta}} = \{x, f^b_{\sigma}(x) - \hat \rho^b_{\beta} \geq 0 \}$
      \EndFor
    \EndFor
	\EndFor
	\State For all $\beta$ and all $\sigma$, compute the volume $\mu^{\sigma}_{\beta}$ of the set $\{x, F^B_{\sigma,\beta}(x) \geq 0 \}$ where $F^B_{\sigma,\beta}(x)= \frac{1}{B}\sum_{b=1}^B (f^b_{\sigma}(x) - \hat \rho^b_{\beta})$
  \State Compute Area under the Mass Volume curve $(\beta, \mu^{\sigma}_{\beta})$ for each $\sigma$: AMV($\sigma$)
  \State $\sigma_{opt} = \argmin_{\sigma \in \Sigma} \text{AMV}(\sigma)$
	\State \Return $\hat G^B_{\alpha}=\{x, F^B_{\sigma_{opt},\alpha}(x) \geq 0\}$ where $F^B_{\sigma_{opt},\alpha}=\frac{1}{B}\sum_{b=1}^B (f^b_{\sigma_{opt}}(x) - \hat \rho^b_{\alpha})$
\end{algorithmic}
\caption{Aggregation of the models learnt on different train/test splits}
\label{OCSVMoffsetVariant}
\end{figure}


\begin{prop}[Nested sets]
Considering several values $0 < \alpha_1 < \dots < \alpha_N < 1$, we can construct nested sets $\hat G^B_{\alpha_1} \subset \dots \subset G^B_{\alpha_N}$.
\end{prop}

\begin{IEEEproof}
For $i \in \{1, \dots, N \}$, let $(f^{b,i}_{\sigma}, \hat \rho^b_i)$, $1 \leq b \leq B$ be the models obtained on the sequence of training and test sets for the mass $\alpha_i$. We have $f^{b,i}_{\sigma} = f^{b,j}_{\sigma}$ for all $i,j \in \{1,\dots, N\}$ as $f^{b,i}_{\sigma}$ only depends on the train and test split. By construction we also have $\hat \rho^b_1 \geq \dots \geq \hat \rho^b_N$ for all $b$. Then for all $b$,
\begin{equation*}
f^{b,1}_{\sigma} - \hat \rho^b_1 \leq \dots \leq f^{b,N}_{\sigma} - \hat \rho^b_N \enspace .
\end{equation*}
By summing
\begin{equation*}
F^{B,1}_{\sigma} \leq \dots \leq F^{B,N}_{\sigma}
\end{equation*}
and if $\hat G^B_{\alpha_i} = \{x, F^{B,i}_{\sigma}(x) \geq 0\}$ then
\begin{equation*}
\hat G^B_{\alpha_1} \subset \dots \subset \hat G^B_{\alpha_N} \enspace .
\end{equation*}
\end{IEEEproof}

\section{Experiments}

For all the experiments we choose $\nu = 1 -\alpha$ for the OCSVM and $\nu=0.4$ for our approach. With our approach $80 \%$ of the data set is used as the training set and the other $20 \%$ as the test set. Unless stated otherwise, $\alpha=0.95$, the Mass Volume curves are made from 10 masses equally spaced between 0.91 and 0.99 and we uniformly sample 10000 points in the smallest hypercube enclosing the data to compute the volumes. All the experimental work was done with Scikit-learn \cite{scikit-learn} using the underlying LIBSVM library \cite{Chang2011}.

\subsection{Simulation with bimodal distribution}
\label{experiments_gm}

\begin{figure}[h!]
\centering
\includegraphics[width=3.5in]{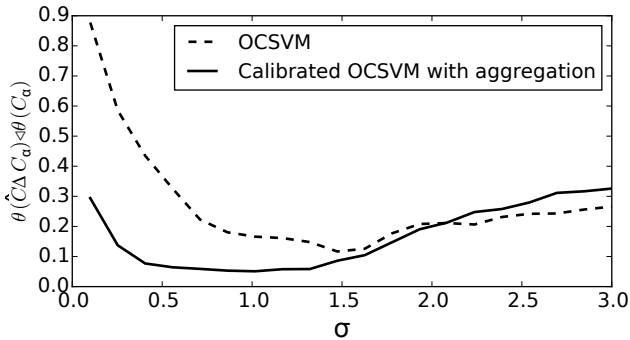}
\caption{Performance as a function of $\sigma$: OCSVM (dashed line) and our approach with an aggregation of 10 models (solid line)}
\label{perf_gm_sigma}
\end{figure}

\begin{figure}[h!]
\centering
\includegraphics[width=3.5in]{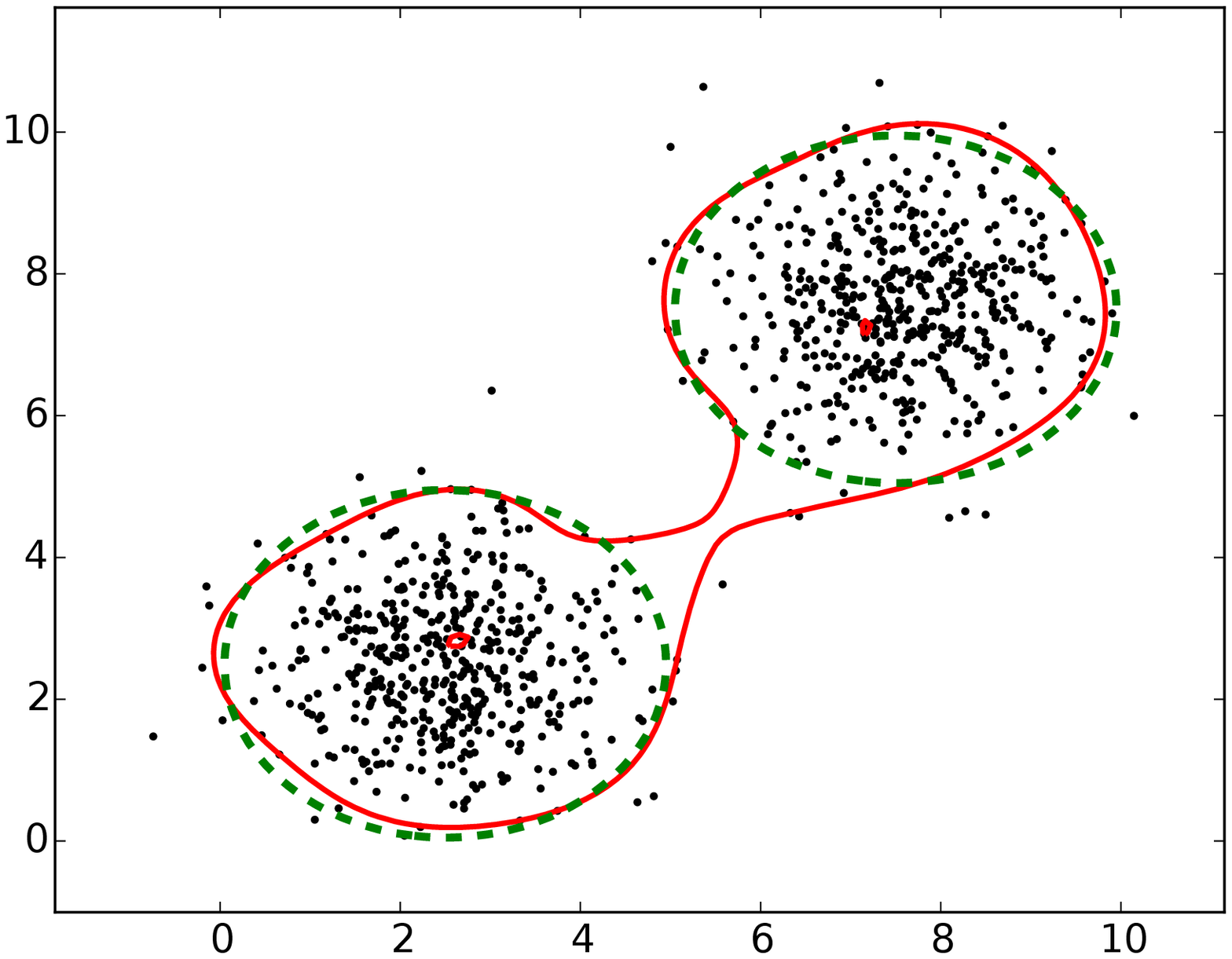}
\caption{In dashed line the true MV set. In solid line the estimated MV set for the best $\sigma$ of the OCSVM with respect to the measure of the symmetric difference between the true and the estimated MV set shown in Figure \ref{perf_gm_sigma}.}
\label{gm_bestsigma}
\end{figure}

\begin{figure}[h!]
\centering
\includegraphics[width=3.5in]{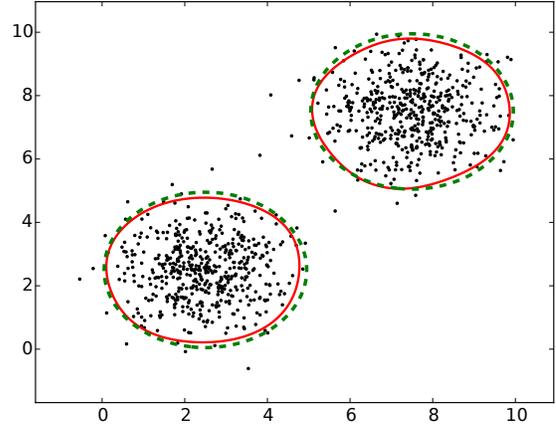}
\caption{In dashed line the true MV set. In solid line the estimated MV set for the best $\sigma$ of our approach with respect to the measure of the symmetric difference between the true and the estimated MV set shown in Figure \ref{perf_gm_sigma}.}
\label{gm_calibrated_best_sigma}
\end{figure}

We sample $n = 1000$ points from a two-dimensional Gaussian mixture of density $h(x) =\frac{1}{2}\mathcal{N}((2.5, 2.5), I)(x) + \frac{1}{2}\mathcal{N}((7.5, 7.5), I)(x)$ where $I$ denotes the identity matrix and $\mathcal{N}(m, \Sigma)(x)$ the density of the Gaussian distribution with mean $m$ and covariance $\Sigma$. We want to estimate the MV set with mass at least $0.95$ from this sample. Knowing the density, we only need the level $\tau_{\alpha}$ such that $P(h(X) \geqslant \tau_{\alpha}) = \alpha$ to know the true MV set $G^*_{\alpha}$. $\tau_{\alpha}$ is the $1-\alpha$ quantile of the distribution of $h(X)$. We estimate such a quantile with 1 million points generated from $h$. To compute the volume of the symmetric difference between the estimated set and the true MV set we sample points uniformly in the hypercube enclosing the data. Our approach is implemented with an aggregation of 10 models. The comparison of the performance as a function of $\sigma$ between the OCSVM and our approach is shown in Figure \ref{perf_gm_sigma}. We observe that the performance of the OCSVM obtained for the best value of $\sigma$, i.e., the value of $\sigma$ minimizing this performance, is worse than the performance reached for a wide range of values of $\sigma$ with our approach. We represent the sets obtained for the values of $\sigma$ giving the best performance for each approach in Figures \ref{gm_bestsigma} and \ref{gm_calibrated_best_sigma}. The solution obtained with our approach is clearly better. Besides, even the solution obtained for the best $\sigma$ of OCSVM tends to overfit (Figure \ref{gm_bestsigma}). \newline

\begin{figure}[h!]
\centering
\includegraphics[width=3.5in]{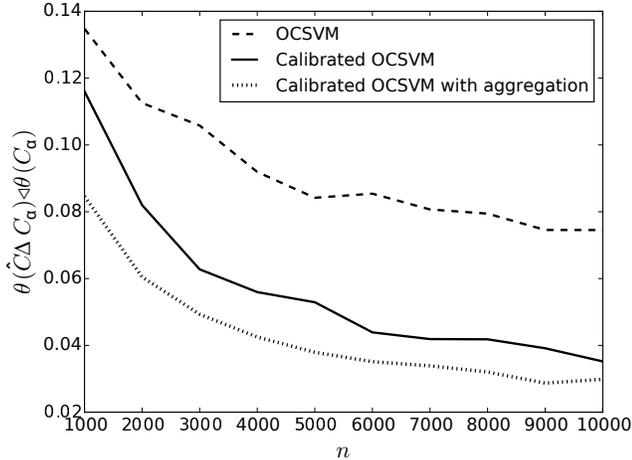}
\caption{Performance as a function of the number of samples $n$. OCSVM (dashed line), our approach without aggregation (solid line) and our approach with an aggregation of 3 models (dotted line).}
\label{comp_perf_n10000_iter100_n_est1}
\end{figure}

In Figure \ref{comp_perf_n10000_iter100_n_est1} we show the evolution of the measure of the symmetric difference between the true and the estimated MV set with mass at least 0.95 as a function of the number of samples. The results are averaged over 100 repetitions. For each sample size, the best $\sigma$ is computed by minimization of the area under the Mass Volume curve for our approach and through minimization of the measure of the symmetric difference for the OCSVM. Again in the case of OCSVM the ground truth is assumed to be known for parameter tuning while our approach automaticaly tune $\sigma$ without the knowledge of the ground truth. Despite this, our approach outperforms the OCSVM when we consider the measure of the symmetric difference between the true and the estimated MV set metric. The approach with aggregation further improves the performance without. \newline

\subsection{Bimodal distribution with outliers}

\begin{figure}[h!]
\centering
\includegraphics[width=3.5in]{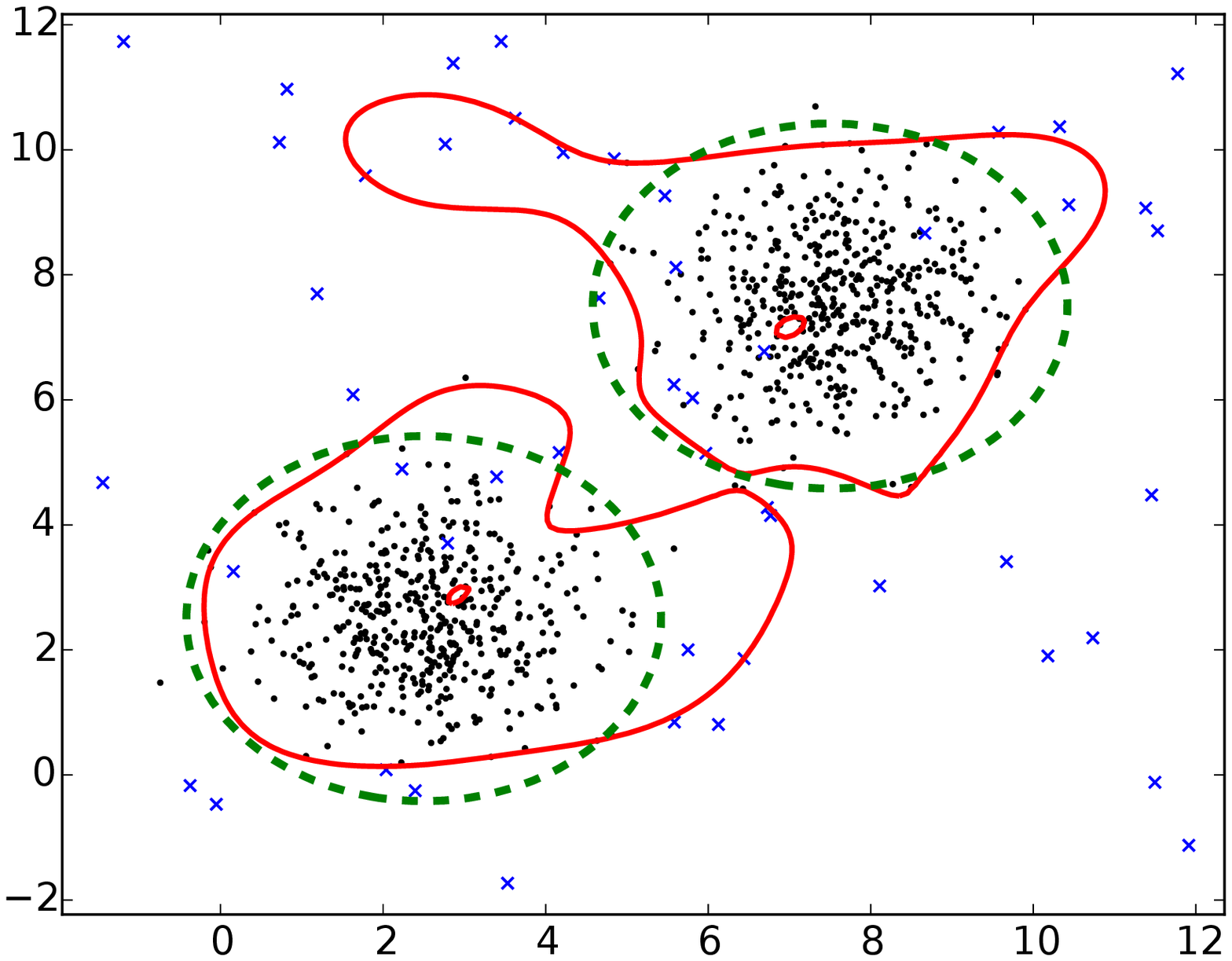}
\caption{In dashed line the true MV set. In solid line the estimated MV set. Outliers are represented by crosses.}
\label{gm_out_best_sigma}
\end{figure}

\begin{figure}[h!]
\centering
\includegraphics[width=3.5in]{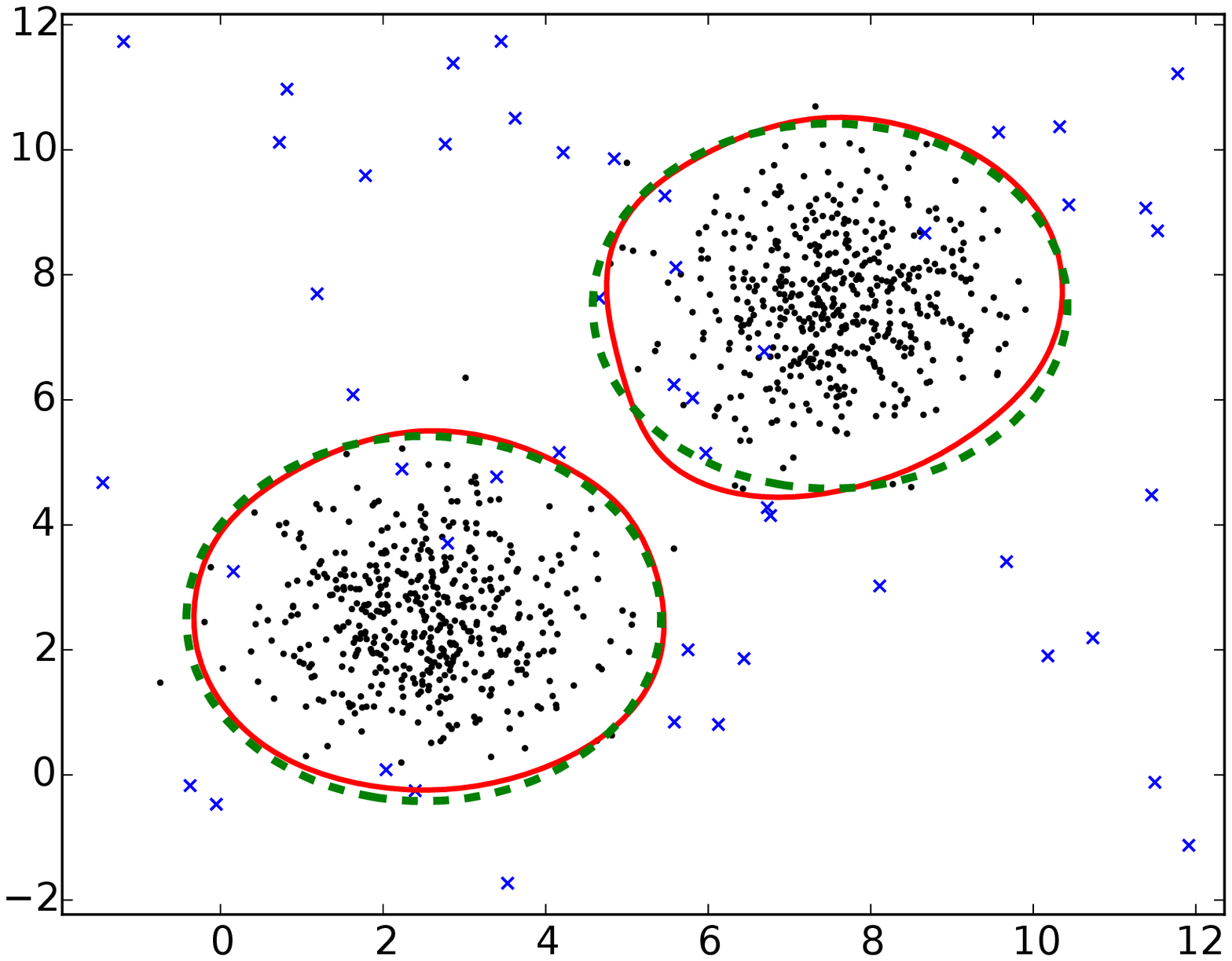}
\caption{In dashed line the true MV set. In solid line the estimated MV set. Outliers are represented by crosses.}
\label{gm_out_calibrated_best_sigma}
\end{figure}

We now considered a two-dimensional Gaussian mixture sample to which we add $5\%$ outliers uniformly distributed over an hypercube enclosing the data. We thus sample $n = 1000$ points from the distribution with density $h(x) = 0.475 \cdot \mathcal{N}((2.5, 2.5), I)(x) + 0.475 \cdot \mathcal{N}((7.5, 7.5), I)(x) + \frac{0.05}{V_C}\mathbbm{1}_C(x)$ where $C = [-2, 12] \times [-2, 12]$ and $V_C$ is the volume of $C$. Knowing the density, we proceed as in section \ref{experiments_gm} to compute the true MV set $G^*_{\alpha}$ of such a distribution. For the OCSVM we choose the value of $\sigma$ minimizing the measure of the symmetric difference between the estimated set and the true MV set. The estimated set is shown in Figure \ref{gm_out_best_sigma}. Our approach is implemented with an aggregation of $10$ models. We consider $20$ values of $\sigma$ equally spaced between $0.01$ and $3$. The best $\sigma$ is obtained by minimization of the area under the Mass Volume curve. The estimated set is shown in Figure \ref{gm_out_calibrated_best_sigma}. This experiment suggests that our approach is more robust to outliers than the OCSVM.

\subsection{Comparison with plug-in approach}

\begin{figure}[h!]
\centering
\includegraphics[width=3.5in]{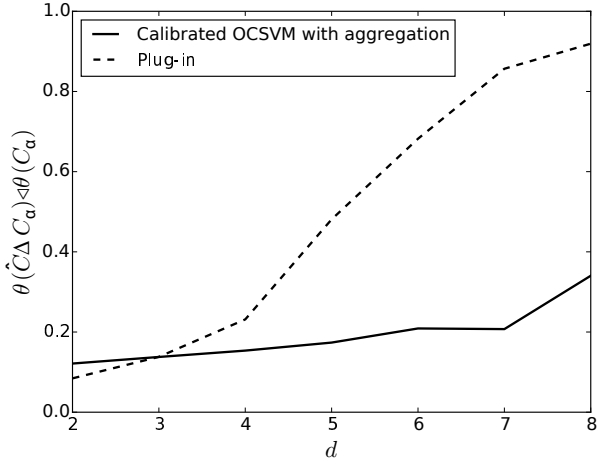}
\caption{Performance as a function of the number of features $d$. Plug-in approach (dashed line) and our approach with an aggregation of 5 models (solid line). Our approach clearly outperforms the plug-in approach as soon as dimension $d$ increases.}
\label{comp_kde_ocsvm}
\end{figure}

In this section we compare the performance of the plug-in approach with our approach with respect to the number of features $d$ for a Gaussian mixture. We recall here that the plug-in approach consists in estimating the underlying density and then thresholding it at the level $\hat \tau_{\alpha}$ such that $P(\{ \hat h_n \geq \hat \tau_{\alpha} \}) = \alpha$. The performance metric used to compare both approach is the measure of the symmetric difference between the true and the estimated MV set with mass at least 0.95. We generate a Gaussian mixture sample of size n=500 with density $h(x) =\frac{1}{2}\mathcal{N}(2.5\cdot\mathbf{1}_d, I_d)(x) + \frac{1}{2}\mathcal{N}(7.5\cdot\mathbf{1}_d, I_d)(x)$, $\mathbf{1}_d$ denoting the vector of $\mathbb{R}^d$ with all its components equal to 1 and $I_d$ denoting the identity matrix of dimension $d$. For the plug-in approach we use a kernel density estimator $\hat h_n$ to estimate $h$ and a bisection search to estimate $\hat \tau_{\alpha}$. The kernel used is the Gaussian kernel with same bandwidth $s$ in all the directions. The bandwidth $s$ is selected through a 4-fold cross validation among 15 values equally spaced between 0.1 and 10. Then we threshold $\hat h_n$ at $\hat \tau_{\alpha}$ such that $P_n(\hat h_n \geq \hat \tau_{\alpha}) = \alpha$ where $P_n$ is the empirical probability measure based on the sample of size $n$. Our approach is performed with an aggregation of $5$ models and the kernel bandwidth is automaticaly selected through minimization of the area under the Mass Volume Curve. In Figure \ref{comp_kde_ocsvm} we show the evolution of the performance for both approach. The results are averaged over 100 repetitions. Even though the performance of both approach is quite similar for $d=2$ and $d=3$, for $d > 3$ we observe that the performance of the plug-in approach deteriorates much more faster than the performance of our approach. We limit this experiment to $d = 8$ because of the difficulty to compute volumes in high dimension.

%
%

\subsection{Two moons data set}
\begin{figure}[h!]
\centering
\includegraphics[width=3.5in]{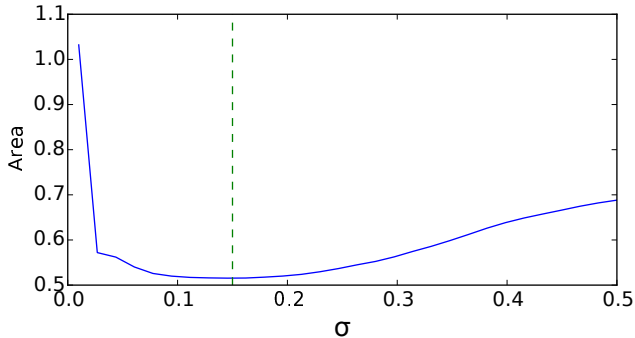}
\caption{Area under the mass volume curve as a function of $\sigma$ for the two moons data set. The minimum is reached at $\sigma =0.15$.}
\label{2moons_amv}
\end{figure}

\begin{figure}[h!]
\centering
\includegraphics[width=3.5in]{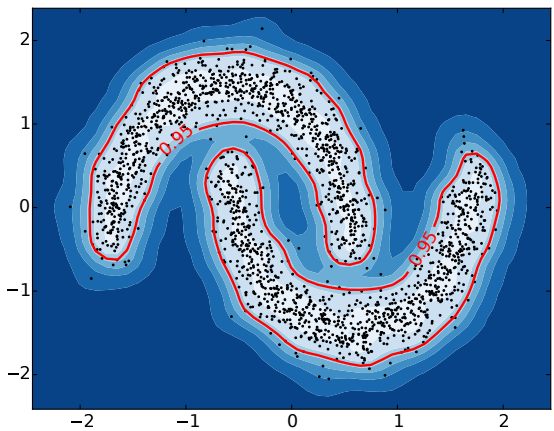}
\caption{Estimated MV set with mass at least 0.95 for a generated two moons data set}
\label{twomoons}
\end{figure}

We generate a two-dimensional two moons data set of size $n=2000$ and try to estimate a MV set with mass at least 0.95. We choose 30 values of $\sigma$ equally spaced between $0.01$ and $0.5$. We average $25$ models based on $25$ train/test random splits of the data set. The best $\sigma$ obtained by minimization of the area under the Mass Volume curve is $\sigma = 0.15$ (see Figure \ref{2moons_amv}). The estimated set is represented in Figure \ref{twomoons}. Its empirical mass on the whole data set is 0.96.

\subsection{Real data set}
\begin{figure}[h!]
\centering
\includegraphics[width=3.5in]{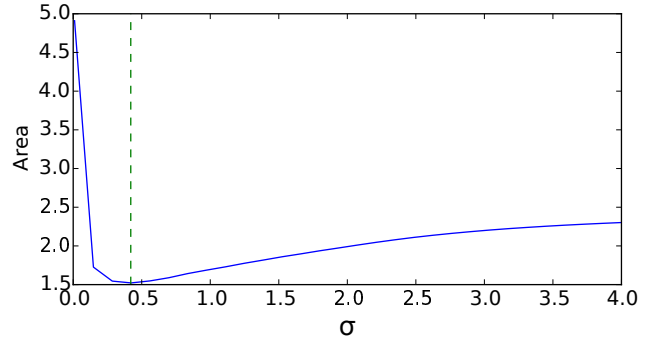}
\caption{Area under the mass volume curve as a function of $\sigma$ for the Boston housing data set. The minimum is reached at $\sigma =0.42$.}
\label{boston_amv}
\end{figure}

\begin{figure}[h!]
\centering
\includegraphics[width=3.7in]{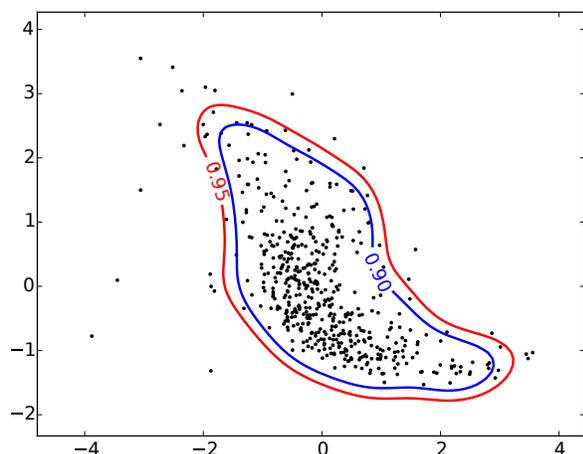}
\caption{MV sets with mass at least 0.90 and 0.95 respectively in blue and in red estimated from the two features, average number of rooms per dwelling (x axis) and percentage lower status of the population (y axis), of the Boston housing data set. The features have been standardized.}
\label{boston_banana}
\end{figure}

We consider here the Boston housing data set \cite{BostonDataset} from the UCI machine learning repository. This data set concerns housing values in suburbs of Boston and consists in $n=506$ samples and $d=14$ features which can be either categorical, integer or real. We only consider two of the features for a better representation of our approach: the average number of rooms per dwelling and the percentage lower status of the population. We first standardize the features, i.e., component wise center and scale to unit variance, and then apply our approach to estimate MV sets. We choose 30 values of $\sigma$ equally spaced between $0.01$ and $4$. We average $25$ models based on $25$ train/test random splits of the data set. The best $\sigma$ obtained by minimization of area under the Mass Volume curve is $\sigma = 0.42$ (see Figure \ref{boston_amv}). The estimated sets are represented in Figure \ref{boston_banana}. The estimated MV set with mass at least 0.90 has an empirical mass of $0.91$ on the whole data set and the estimated MV set with mass at least 0.95 has en empirical mass of $0.95$. We observe that the estimated sets are nested.

\section{Conclusion}

This paper presents a new approach to estimate MV sets using the OCSVM algorithm. Results show that it outperforms the standard way to use the OCSVM. Our approach is based on the calibration of the offset of the solution function to obtain the desired probability mass on a test set. It allows to compute nested set estimates without the need to add any condition ensuring this property and consider several regularization parameters. Moreover it provides a scoring rule for samples located in the tail of the underlying distribution. The computed Mass Volume curve allows to assess the performance of the approach and to select the kernel bandwidth automatically. Our solution inherits the sparsity of the OCSVM which is a computational advantage over kernel smoothing.

The kernel bandwidth selection requires to compute the volume of the estimated set which suffers from the curse of dimensionality. This issue is still an open research area. Sampling more precisely in the region where the data lives instead of sampling in the hypercube enclosing the data is a possible approach to scale to higher dimensions.






%

\IEEEtriggeratref{10}
\IEEEtriggercmd{\enlargethispage{-2.10in}}

\bibliographystyle{IEEEtran}
\bibliography{IEEEabrv,biblio}
%
%

\end{document}